\documentclass[10pt,conference]{IEEEtran}
\IEEEoverridecommandlockouts
\usepackage{cite}
\usepackage{amsmath,amssymb,amsfonts}
\usepackage{algorithmic}
\usepackage{graphicx}
\usepackage{textcomp}
\usepackage{xcolor}
\usepackage{colortbl}
\usepackage{multirow}
\usepackage[export]{adjustbox}
\usepackage{float}
\usepackage{subfigure}
\usepackage{hyperref}

\def\BibTeX{{\rm B\kern-.05em{\sc i\kern-.025em b}\kern-.08em
    T\kern-.1667em\lower.7ex\hbox{E}\kern-.125emX}}
\begin{document}

\title{Quality In / Quality Out: Data quality more relevant than model choice in anomaly detection with the UGR'16

\thanks{This work was supported by the Agencia Estatal de Investigación in Spain, grant  No PID2020-113462RB-I00, and the European Union’s Horizon 2020 research and innovation programme under the Marie Skłodowska-Curie grant agreement No 893146. IEEE policy provides that authors are free to follow funder public access mandates to post accepted papers in repositories. When posting in a repository, the IEEE embargo period is 24 months. However, IEEE recognizes that posting requirements and embargo periods vary by funder. IEEE authors may comply with requirements to deposit their accepted papers in a repository per funder requirements where the embargo is less than 24 months. Information on specific funder requirements can be found at  \url{https://open.ieee.org/for-authors/funders/}.}
}

\author{
\IEEEauthorblockN{José Camacho}
\IEEEauthorblockA{\textit{Dept. Signal Theory, Networking and } \\
\textit{Communication, University of Granada}\\
Granada, Spain \\
josecamacho@ugr.es}
\and
\IEEEauthorblockN{Katarzyna Wasielewska}
\IEEEauthorblockA{\textit{Dept. Signal Theory, Networking and } \\
\textit{Communication, University of Granada}\\
Granada, Spain \\
k.wasielewska@ugr.es}
\and
\IEEEauthorblockN{Pablo Espinosa}
\IEEEauthorblockA{\textit{Dept. Signal Theory, Networking and } \\
\textit{Communication, University of Granada}\\
Granada, Spain \\
pabloespinosa@correo.ugr.es}
\and
\IEEEauthorblockN{Marta Fuentes-Garc{\'i}a}
\IEEEauthorblockA{COSCYBER\\
FIDESOL\\
Granada, Spain \\
mfuentes@fidesol.org}}

\maketitle

\begin{abstract}
Autonomous or self-driving networks are expected to provide a solution to the myriad of extremely demanding new applications with minimal human supervision. For this purpose, the community relies on the development of new Machine Learning (ML) models and techniques. 
However, ML can only be as good as the data it is fitted with, and data quality is an elusive concept difficult to assess. In this paper, we show that relatively minor modifications on a benchmark dataset (UGR'16, a flow-based real-traffic dataset for anomaly detection) cause significantly more impact on model performance than the specific ML technique considered. We also show that the measured model performance is uncertain, as a result of labelling inaccuracies. Our findings 
illustrate that the widely adopted approach of comparing a set of models in terms of performance results (e.g., in terms of accuracy or ROC curves) may lead to incorrect conclusions when done without a proper understanding of dataset biases and sensitivity. We contribute a methodology to interpret a model response that can be useful for this understanding.  
\end{abstract}


\section{Introduction}

There is an increasing interest in the development of new machine learning (ML) methods to improve the performance of communication networks~\cite{kalmbach2018empowering}. ML tools can only be as good as the data they are trained on, reason why we need high-quality datasets \cite{Caviglione2021}\cite{Sarker2020}. However, while the process of model optimization and the development of new ML methods have received the full attention of the community, techniques to assess data quality are scarce and often ignored \cite{camacho2022dataset}.

In this paper, we show that the impact of minor data modifications prior to modelling with ML can be indeed more relevant than the specific ML method used. These modifications include mild changes on how traffic features were computed, whether or not data was anonymized, and the set of observations that were considered for model fitting and testing. This case study illustrates that the research community needs to look more into data quality assessment and optimization.

Our main contribibutions are:

\begin{itemize}
    \item We derive four variants of a benchmark dataset in network anomaly detection, by applying minor differences in the data treatment. We perform anomaly detection using these variants with two very different ML methodologies, finding negligible differences in performance between the ML variants but significant differences among the dataset variants.
    
    \item We develop an analysis methodology to investigate the root causes of the performance differences found. Applying this methodology to the case study provides a full understanding of the differences, which allows us to obtain a better picture of when these are relevant and/or when they are due to labelling inaccuracies (in particular, unlabelled anomalies). 
    
    
\end{itemize}

The paper is organized as follows. Section \ref{sec:MM} introduces the case study under analysis, the preprocessing and data selection steps, the ML methods considered and our approach to interpret model performance. Section \ref{sec:experimental} presents the experimental results and Section \ref{sec:conclusions} draws the conclusions.

\section{Materials and Methods}
\label{sec:MM}

The UGR'16 dataset~\cite{macia2018ugr}\footnote{Dataset available online at \url{https://nesg.ugr.es/nesg-UGR'16/}} was captured from a real network of a tier 3 Internet Server Provider (ISP). The data collection was carried out with Netflow between March and June of 2016 under Normal Operation Conditions (NOCs), meaning that the network was used normally by the ISP clients. This allowed to model and study the normal behavior of the network, and to unveil certain anomalies such as SPAM campaigns. The flows of the dataset were labelled indicating if they were "background" (regarded as legitimate flows), or "anomalies" (identified as non-legitimate flows). In addition, another capture was made between July and August of 2016, including some controlled attacks that were launched to obtain a test dataset for validation of anomaly detection algorithms. The type of attacks were \textit{Denial of Service} (DOS), \textit{port scanning} in two modalities: either from one attacking machine to one victim machine (SCAN11) or from four attacking machines to four victim machines (SCAN12), and \textit{botnet traffic} (NERISBOTNET). As of today, the UGR'16 has been referenced in more than 160 research papers (according to Google Scholar) and it can be considered a benchmark in the research of anomaly detection in real traffic data for cybersecurity.

We made use of the \textit{feature-as-a-counter} (FaaC) approach in this work  \cite{Camacho2014}. Using this approach, we performed anomaly detection at 1 minute intervals rather than at flow level. A total of 134 features were extracted per interval. The process of feature extraction was based on two steps: i) binary files were transformed to flow-level csv files with the nfdump tool, and ii) csv files were transformed to feature vectors with the FCParser \cite{CAMACHO2019101603}. 

\begin{table}[t]
	\caption{UGR'16 dataset variants.}
	\label{tab:variants}
	\centerline{
		\small{
			\begin{tabular}{l c c c}
				\hline \\[-1.5ex]
				{\bf Label} & {\bf Training } & {\bf Type of flows} & {\bf Anonymized flows} \\[0.5ex]
				\hline \\[-1.5ex]
				UGR'16v1 & March to June & Unidirectional & No \\
				UGR'16v2 & March to May & Unidirectional & No \\
				UGR'16v3 & March to May & Bidirectional & Yes \\
				UGR'16v4 & March to May & Unidirectional & Yes 
				\\[0.5ex]
				\hline
			\end{tabular}
	}}
	
\end{table}

We considered four variants of the UGR'16, described in Table \ref{tab:variants}: 

\begin{itemize}
    \item The first variant (UGR'16v1) included the original (non-anonymized) Netflow logs for the entire NOC period (from March to June). This corresponds to the same data used in previous works \cite{CAMACHO2019101603}. 
    \item Fuentes \cite{fuentes2021multivariate} found that the training data corresponding to June included real anomalies that hamper the ability of detection of the botnet attack in the test set. Leveraging this finding, we considered a second version (UGR'16v2) in which the training data corresponds only to the period from March to May. 
    \item Both previous versions (UGR'16v1 and UGR'16v2) considered unidirectional Netflow flows, which may complicate the interpretation of the results. For this reason, we decided to repeat the feature generation process using bidirectional flows (in nfdump), in this case considering the anonymized flows available online. This is the third version of the dataset (UGR'16v3), and it shared with the second version that June is not included in the training data. 
    \item Finally, and to distinguish the influence of anonymization from the use of bidirectional or unidirectional flows, we considered a last version (UGR'16v4) equivalent to version 3 but with unidirectional flows. 
\end{itemize}

The consideration of the previously described four versions of UGR'16 allowed us to determine the impact of some data preprocessing steps on the model quality for anomaly detection, in particular:

\begin{itemize}
    \item The selection of the set of training data (by comparing performance results between UGR'16v1 and UGR'16v2).
    \item The effect of bi- or uni-directional flows (by comparing performance results between UGR'16v3 and UGR'16v4).
    \item The effect of anonymization (by comparing performance results between UGR'16v2 and UGR'16v4).
    
\end{itemize}

To compare the influence of data preprocessing methods in the anomaly detection performance against the influence of the specific ML methods used, we considered two very different tools: the Multivariate Statistical Network Monitoring (MSNM) \cite{MSNM2015} and the one-class support vector machine (OCSVM)  \cite{Scholkopf2000,Scholkopf2001} based on radial basis functions (RBF), the most extended kernel choice. The former is a linear multivariate approach, and therefore it is specially suited to handle the highly multivariate nature of the FaaC features. The latter is a non-linear tool, and therefore has the advantage to model non-linear behaviour in the model of normal traffic. Thus, both methods have very different features that could, in principle, affect performance in a significant way.



To test the anomaly detection performance with the different data and model variants, we computed the false positive rate (FPR) and true positive rate (TPR) in the labeled test set, and in turn the Receiver Operating Characteristic (ROC) curves, that show the evolution of the TPR versus the FPR for different values of the anomaly detection threshold. We selected this option since in the context of network security, maintaining the balance between TP and FP is relevant in practice \cite{alpcan2010network,collins2014network}. A practical way to compare several ROC curves is with the Area Under the Curve (AUC), a scalar that quantifies the quality of the anomaly detector. An anomaly detector should present an AUC as close to 1 as possible, while an AUC {around} 0.5 corresponds to a random classifier.  

To shade light on the model performance differences when using different dataset versions, we used the Univariate-Squared (U-Squared) statistic \cite{FUENTESGARCIA2018194}. The U-Squared statistic provides a discriminative pattern for the attack in comparison to the reference. In our case, this reference is represented by any of the versions of the UGR'16. This pattern can be leveraged to determine whether the reference dataset is of good quality to train anomaly detection models able to detect the attack or not. Given a specific attack/anomaly,
the U-Square provides us with a subset of relevant features, and then we can proceed using statistical means to analyze if those features have good detection capability for the attack. We will show that this approach can provide a full understanding of the performance differences between dataset variants in our case study.

\section{Experiments and results}
\label{sec:experimental}

\begin{figure}[!h]
	\centering
	\subfigure[]{
		\includegraphics[width=0.7\linewidth]{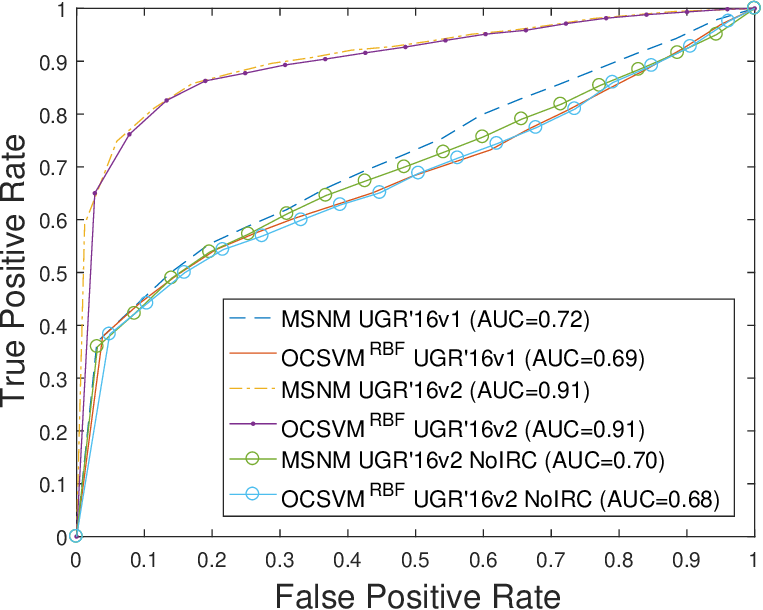}
	}
	\subfigure[]{
		\includegraphics[width=0.7\linewidth]{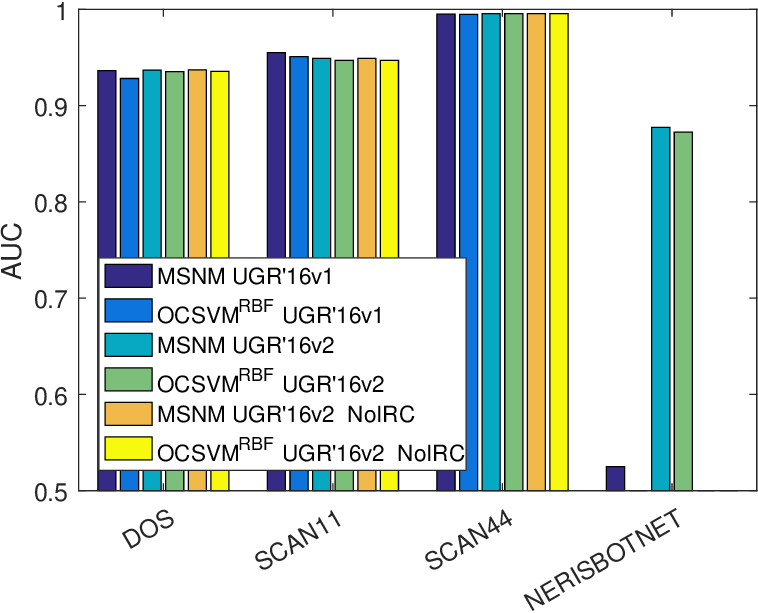}
	}
	\caption{ROC curve (a) and attack-type based AUC results (b) for the data parsed from original unidirectional flows in UGR'16v1 and UGR'16v2, and for a variant of the latter with no IRC features (UGR'16v2 NoIRC).} \label{ROC_ori}
\end{figure}

Fig. \ref{ROC_ori} shows the comparison of the two anomaly detectors (MSNM and OCSVM) when trained with the datasets UGR'16v1 and UGR'16v2, and with a sub-version of UGR'16v2 (UGR'16v2 NoIRC) that will be discussed later. Fig. \ref{ROC_ori}(a) presents the general ROC curves, obtained for the four types of attacks, and Fig. \ref{ROC_ori}(b) represents the AUCs per attack type. Performance differences between the two anomaly detectors are minor in all cases. However, there is a huge difference with respect to including June in the training data (UGR'16v1) or not including it (UGR'16v2). This difference can be mapped to one specific attack type, the NERISBOTNET. We hypothesize that this difference is mainly caused by the anomaly detected in the background traffic of June, related to suspicious activity in MIRC \cite{fuentes2021multivariate}. 

\begin{figure}[!h]
	\centering
	\subfigure[]{
		\includegraphics[width=0.45\linewidth]{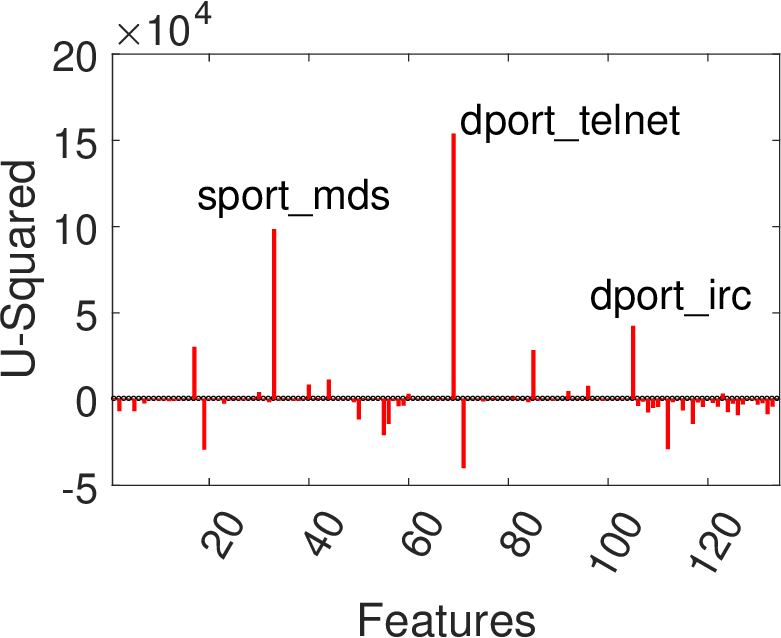}
	}
	\subfigure[]{
		\includegraphics[width=0.45\linewidth]{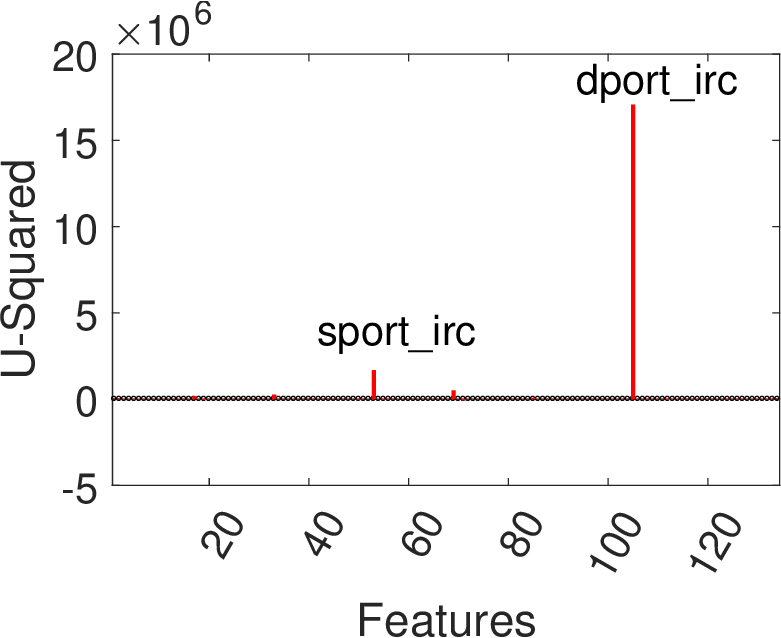}
	}
	\caption{Comparison of U-Squared statistics for the NERISBOTNET attack using as a reference UGR'16v1 (a) and UGR'16v2 (b).} \label{oMEDA}
\end{figure}

To check our hypothesis, we compute the U-Squared statistic for the observations in the test set that contain flows of the NERISBOTNET attack, and using as a reference UGR'16v1 and UGR'16v2, respectively. This is shown in Fig. \ref{oMEDA}. When using UGR'16v1 as a reference (Fig. \ref{oMEDA}(a)), we find that the NERISBOTNET attack is mainly characterized by an excess in 3 out of the 134 features: $sport\_mds$, $dport\_telnet$ and $dport\_irc$. This suggests that the number of flows with source port MDS, with destination port TELNET and with destination port IRC are generally higher in observations where NERISBOTNET attacks are taking place. However, when we use UGR'16v2 as a reference (Fig. \ref{oMEDA}(b)), the NERISBOTNET attack is mainly characterized by the amount of flows to or from the IRC port. This difference between the U-Squared patterns found with the two reference datasets implies that ML models trained from them will have different means to detect the NERISBOTNET attack. These differences affect performance, as seen in the AUC results.

\begin{figure}[!t]
	\centering
	\subfigure[Ttest $p_-value = 0.47$]{
		\includegraphics[width=0.4\linewidth]{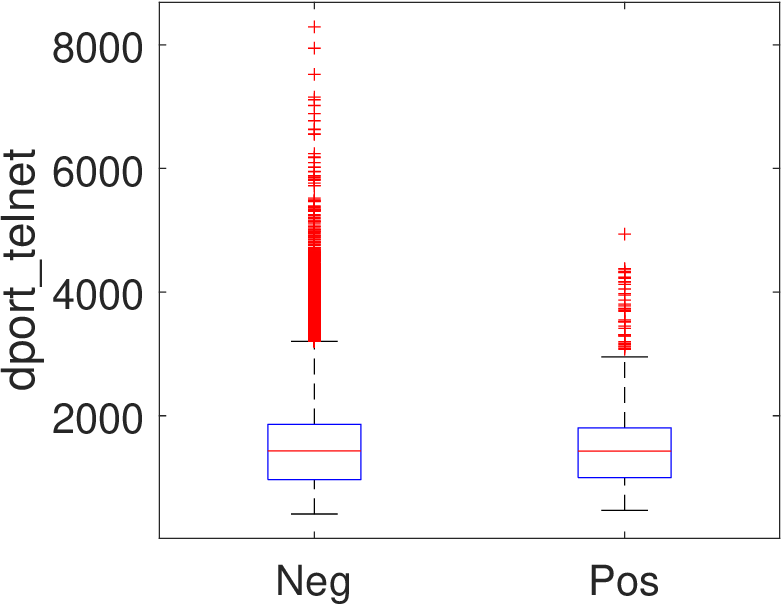}
	}
\subfigure[Ttest $\mathbf{p_-value < 0.01}$]{
		\includegraphics[width=0.4\linewidth]{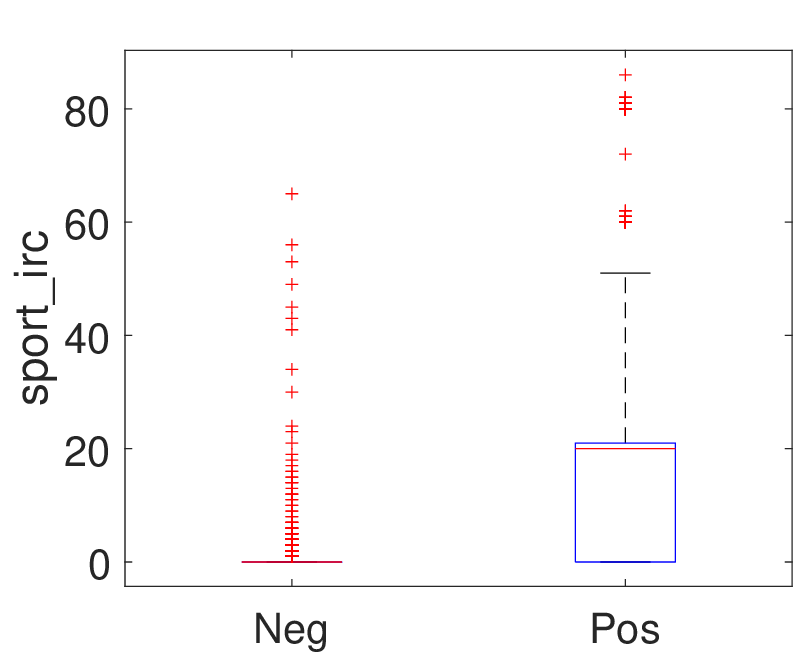}
	}
	\caption{Boxplots of selected features in background traffic (Negative) versus NERISBOTNET traffic (Positive).} \label{Boxs}
\end{figure}

Fig. \ref{Boxs} presents boxplots to compare the distribution, in the test set, of the normal vs the NERISBOTNET observations for a set of selected features, previously highlighted by the U-Squared. We also include the result of a t-test to check whether there is statistical evidence that the NERISBOTNET attack does present higher content in the corresponding feature. Feature $dport\_telnet$, highlighted when UGR'16v1 is the reference, does not show statistical significant differences between normal and NERISBOTNET observations. Clearly, including the anomaly in June as "normal data" makes the detectors to incorporate this type of activity in the normality model, and therefore prevents them to detect it in future traffic. 
Therefore, this feature (and in general UGR'16v1) will allow a low detection ability of the attack, regardless the ML method used. Contrarily, all IRC features (only $sport\_irc$ shown, Fig. \ref{Boxs}(b)) do show statistical significant differences. Therefore, we can conclude that models trained with UGR'16v2 will detect the presence of NERISBOTNET attacks as significant changes in the IRC features, and will yield a high detection ability. This conclusion is further supported by the fact that if we train the models with UGR'16v2, but we delete  the IRC features $sport\_irc$ and $dport\_irc$ from the data, the detection of NERISBOTNET is poor, as illustrated in  Fig. \ref{ROC_ori} with the results associated to the label "UGR'16v2 NoIRC". 

This example illustrate the well-known fact that when the training data is contaminated with anomalies, ML methods decrease performance. Unfortunately, detecting such anomalies in real-life data has deserved little attention in the community but can be principal in the context of autonomous networks. In this real example, the proper selection of observations (and features) was by far more relevant than the choice of the ML method employed.

\begin{figure}[!t]
	\centering
		\includegraphics[width=0.7\linewidth]{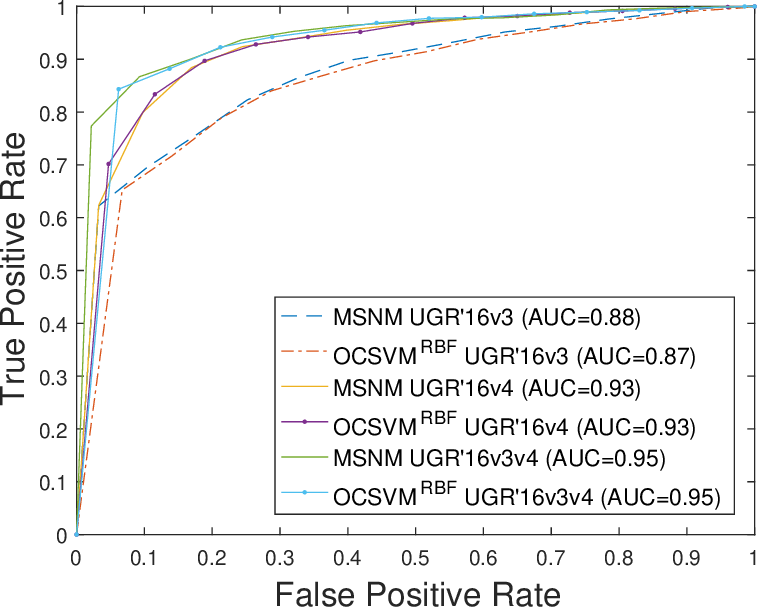}
	
	\caption{ROC curve for the data parsed from anonymized bidirectional (UGR'16v3) and unidirectional (UGR'16v4) flows, and a combination of both (UGR'16v3v4).} \label{ROC_bid}
\end{figure}

\begin{figure}[!t]
	\centering
	\subfigure[Ttest $\mathbf{p_-value < 0.01}$]{
		\includegraphics[width=0.4\linewidth]{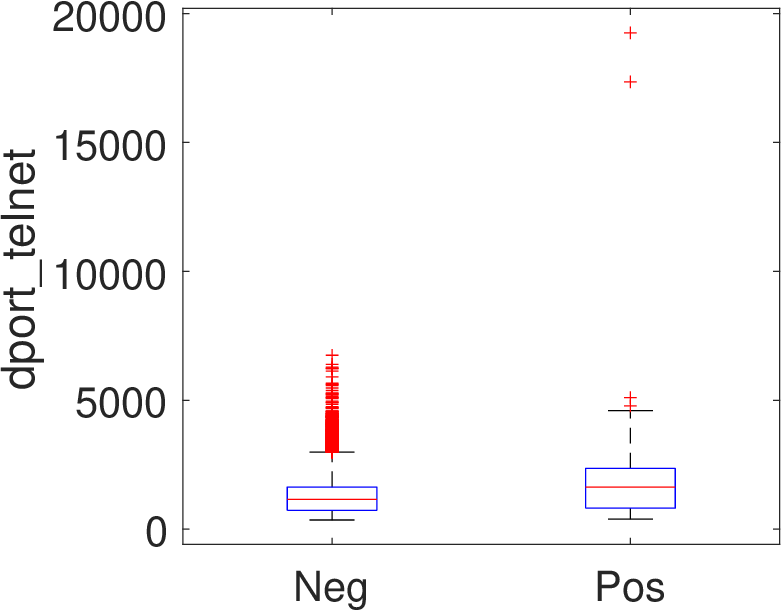}
	}
	\subfigure[Ttest $\mathbf{p_-value < 0.01}$]{
		\includegraphics[width=0.4\linewidth]{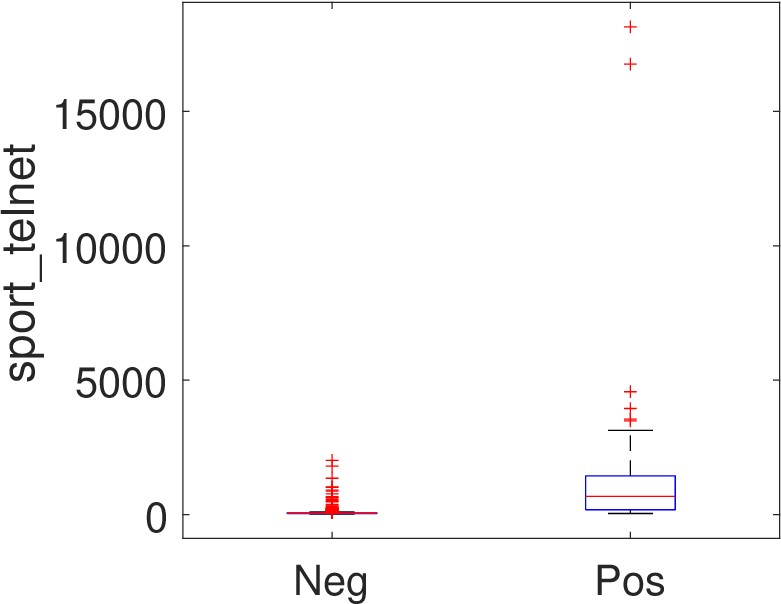}
	}
	\caption{Boxplot in background traffic (Negative) versus DOS traffic (Positive) of dport\_telnet in UGR'16v3 (a) and of sport\_telnet in UGR'16v4 (b).} \label{Boxs4}
\end{figure}

Fig. \ref{ROC_bid} presents the performance results of the anomaly detectors in UGR'16v3 and UGR'16v4, and a combination of both datasets that will be discussed later. In all situations, the differences between the two detectors, MSNM and OCSVM, is again negligible. Performance differences are observed between the use of bidirectional and unidirectional flows, in favour of the latter. In this case, this difference is mainly mapped to the DOS attacks (not shown). Therefore, like in the previous comparison, relatively minor decisions on data preparation (in this case whether or not use an nfdump flag during flows parsing) impact more in the performance than the choice of the ML tool. To shade some light into the observed differences in the detection of DOS attacks, we computed the U-Squared for the observations including  DOS attacks using UGR'16v3 and UGR'16v4 as references (not shown). Again, we find different patterns of characterization depending on the reference dataset. Using bidirectional flows, the DOS attacks are characterized by flows with destination ports HTTP and TELNET. Statistical significant differences between test normal observations and those containing DOS attacks confirm this characterization (see Fig. \ref{Boxs4}(a)). When we use unidirection flows (UGR'16v4), the DOS attacks are only characterized by the activity in the TELNET source port. Fig. \ref{Boxs4}(b) shows this characterization is not only statistically significant but also of high quality: the activity of TELNET source port in normal observations is almost null. This is the explanation for the higher performance of anomaly detection models when using unidirectional flows in DOS attacks. 

It should be noted that bidirectional flows are indeed slightly better in the detection of NERISBOTNET than unidirectional flows (not shown), what suggests that the best detection performance in this case is attack specific. Given that the convenience on the use of unidirectional or birectional flows is attack specific, we can always combine both set of features in a single dataset with double number (268) of  features. We name such dataset UGR'16v3v4. When we do so, the performance is optimized in general terms, as shown in Fig.  \ref{ROC_bid}. 

UGR'16v4 represents the anonymized version of UGR'16v2. Performance results for UGR'16v4 are slightly better than to those in UGR'16v2 (compare Figs. \ref{ROC_ori} and \ref{ROC_bid}), but differences are not robust to minor modifications in the analyses (like  minor modifications in the training observations).   

Finally, we used the same general interpretation approach based on U-Squared, boxplots and t-tests to evaluate seemingly brackground observations in the test set with a high anomaly score in some of the models. We found that these observations indeed followed an attack pattern that was mislabelled. To some extent, this is a similar problem to the one in Fig. \ref{ROC_ori} with the unlabelled anomaly in June. In this case, however, mislabelling in the test dataset affects the reliability of the ROC/AUC.

\section{Conclusions}
\label{sec:conclusions}

In this paper, we present a number of anomaly detection experiments in a real network dataset, the UGR'16, which can be regarded as a benchmark. We show that minor data decisions have a major influence on the performance result and that labelling errors can severely affect the conclusions driven from the benchmark. This illustrate that our community should look more into (automatic) data quality assessment. This is especially important in the context of autonomous networks, where the Machine Learning workflow is expected to have little or none human supervision. In general, performance results should not be provided without a proper understanding of sources of bias and sensitivity in a dataset, but this understanding is challenging and often overlooked. In this paper, we contribute an approach to understand model response in connection with training data and data labelling. Future work may extend this methodology in the automatic assessment of data quality.


This work was supported by the Agencia Estatal de Investigaci´on in Spain, MCIN/AEI/ 10.13039/501100011033, grant No PID2020-113462RB-I00., and by the European Union’s Horizon 2020 research and innovation programme under the Marie Skłodowska-Curie grant agreement No 893146.

\bibliographystyle{IEEEtran}
\bibliography{3way}

\begin{thebibliography}{10}
\providecommand{\url}[1]{#1}
\csname url@samestyle\endcsname
\providecommand{\newblock}{\relax}
\providecommand{\bibinfo}[2]{#2}
\providecommand{\BIBentrySTDinterwordspacing}{\spaceskip=0pt\relax}
\providecommand{\BIBentryALTinterwordstretchfactor}{4}
\providecommand{\BIBentryALTinterwordspacing}{\spaceskip=\fontdimen2\font plus
\BIBentryALTinterwordstretchfactor\fontdimen3\font minus
  \fontdimen4\font\relax}
\providecommand{\BIBforeignlanguage}[2]{{%
\expandafter\ifx\csname l@#1\endcsname\relax
\typeout{** WARNING: IEEEtran.bst: No hyphenation pattern has been}%
\typeout{** loaded for the language `#1'. Using the pattern for}%
\typeout{** the default language instead.}%
\else
\language=\csname l@#1\endcsname
\fi
#2}}
\providecommand{\BIBdecl}{\relax}
\BIBdecl

\bibitem{kalmbach2018empowering}
P.~Kalmbach, J.~Zerwas, P.~Babarczi, A.~Blenk, W.~Kellerer, and S.~Schmid,
  ``Empowering self-driving networks,'' in \emph{Proceedings of the afternoon
  workshop on self-driving networks}, 2018, pp. 8--14.

\bibitem{Caviglione2021}
L.~Caviglione, M.~Choraś, I.~Corona, A.~Janicki, W.~Mazurczyk, M.~Pawlicki,
  and K.~Wasielewska, ``Tight arms race: Overview of current malware threats
  and trends in their detection,'' \emph{IEEE Access}, vol.~9, pp. 5371--5396,
  2021.

\bibitem{Sarker2020}
I.~Sarker, A.~S.~M. Kayes, S.~Badsha, H.~Alqahtani, P.~Watters, and A.~Ng,
  ``Cybersecurity data science: an overview from machine learning
  perspective,'' \emph{Journal of Big Data}, vol.~7, 07 2020.

\bibitem{camacho2022dataset}
J.~Camacho and K.~Wasielewska, ``Dataset quality assessment in autonomous
  networks with permutation testing,'' in \emph{NOMS 2022-2022 IEEE/IFIP
  Network Operations and Management Symposium}.\hskip 1em plus 0.5em minus
  0.4em\relax IEEE, 2022, pp. 1--4.

\bibitem{macia2018ugr}
G.~Maci{\'a}-Fern{\'a}ndez, J.~Camacho, R.~Mag{\'a}n-Carri{\'o}n,
  P.~Garc{\'\i}a-Teodoro, and R.~Ther{\'o}n, ``{UGR‘16}: A new dataset for
  the evaluation of cyclostationarity-based network {IDS}s,'' \emph{Computers
  \& Security}, vol.~73, pp. 411--424, 2018.

\bibitem{Camacho2014}
J.~Camacho, G.~Maci\'a-Fern\'andez, J.~D\'iaz-Verdejo, and P.~Garc\'ia-Teodoro,
  ``{Tackling the big data 4 vs for anomaly detection},'' \emph{Proceedings of
  the IEEE INFOCOM}, no.~1, pp. 500--505, 2014.

\bibitem{CAMACHO2019101603}
\BIBentryALTinterwordspacing
J.~Camacho, J.~M. García-Giménez, N.~M. Fuentes-García, and
  G.~Maciá-Fernández, ``Multivariate big data analysis for intrusion
  detection: 5 steps from the haystack to the needle,'' \emph{Computers \&
  Security}, vol.~87, p. 101603, 2019. [Online]. Available:
  \url{https://www.sciencedirect.com/science/article/pii/S0167404818307909}
\BIBentrySTDinterwordspacing

\bibitem{fuentes2021multivariate}
N.~M. Fuentes~Garc{\'\i}a \emph{et~al.}, ``Multivariate statistical network
  monitoring for network security based on principal component analysis,''
  2021.

\bibitem{MSNM2015}
\BIBentryALTinterwordspacing
J.~Camacho, A.~P\'erez-Villegas, P.~Garc\'ia-Teodoro, and
  G.~Maci\'a-Fern\'andez, ``{PCA}-based multivariate statistical network
  monitoring for anomaly detection,'' \emph{Computers \& Security}, vol.~59,
  pp. 118--137, June 2016. [Online]. Available:
  \url{http://www.sciencedirect.com/science/article/pii/S0167404816300116}
\BIBentrySTDinterwordspacing

\bibitem{Scholkopf2000}
B.~Sch\"{o}lkopf, A.~J. Smola, R.~C. Williamson, and P.~L. Bartlett, ``{New
  Support Vector Algorithms},'' \emph{Neural computation}, vol.~12, no.~5, pp.
  1207--1245, 2000.

\bibitem{Scholkopf2001}
\BIBentryALTinterwordspacing
B.~Sch{\"{o}}lkopf, J.~C. Platt, J.~Shawe-Taylor, A.~J. Smola, and R.~C.
  Williamson, ``{Estimating the Support of a High-Dimensional Distribution},''
  \emph{Neural Computation}, vol.~13, no.~7, pp. 1443--1471, 2001. [Online].
  Available:
  \url{http://www.mitpressjournals.org/doi/abs/10.1162/089976601750264965}
\BIBentrySTDinterwordspacing

\bibitem{alpcan2010network}
T.~Alpcan and T.~Ba{\c{s}}ar, \emph{Network security: A decision and
  game-theoretic approach}.\hskip 1em plus 0.5em minus 0.4em\relax Cambridge
  University Press, 2010.

\bibitem{collins2014network}
M.~Collins and M.~S. Collins, \emph{Network security through data analysis:
  building situational awareness}.\hskip 1em plus 0.5em minus 0.4em\relax "
  O'Reilly Media, Inc.", 2014.

\bibitem{FUENTESGARCIA2018194}
\BIBentryALTinterwordspacing
M.~Fuentes-García, G.~Maciá-Fernández, and J.~Camacho, ``Evaluation of
  diagnosis methods in pca-based multivariate statistical process control,''
  \emph{Chemometrics and Intelligent Laboratory Systems}, vol. 172, pp. 194 --
  210, 2018. [Online]. Available:
  \url{http://www.sciencedirect.com/science/article/pii/S0169743917302046}
\BIBentrySTDinterwordspacing

\end{thebibliography}

\vspace{12pt}

\end{document}